\documentclass[journal]{IEEEtran}
\IEEEoverridecommandlockouts

\usepackage{stfloats}

\usepackage{amsmath,amssymb,amsfonts}
\usepackage{graphicx}
\usepackage{booktabs}
\usepackage{algorithm}
\usepackage{cite}

\usepackage[percent]{overpic}
\usepackage{xcolor} 

\usepackage{url}

\begin{document}
\bstctlcite{IEEEexample:BSTcontrol}

\title{Enhanced Protein Intrinsic Disorder Prediction Through Dual-View Multiscale Features and Multi-objective Evolutionary Algorithm
}

\author{Shaokuan Wang, Pengshan Cui, Yining Qian, An-Yang Lu, and Xianpeng Wang%
\thanks{This work has been submitted to the IEEE for possible publication. Copyright may be transferred without notice, after which this version may no longer be accessible.}%
\thanks{Shaokuan Wang, Pengshan Cui, and An-Yang Lu are with the College of Information Science and Engineering, Northeastern University, Shenyang 110819, China (e-mail: wangsk@mails.neu.edu.cn; ccuipengshan@163.com; luanyang@mail.neu.edu.cn).}%
\thanks{Yining Qian is with the School of Computer Science and Engineering, Northeastern University, Shenyang 110819, China (e-mail: qianyiningning@126.com). Corresponding author.}%
\thanks{Xianpeng Wang is with the Key Laboratory of Data Analytics and Optimization for Smart Industry, Ministry of Education, Northeastern University, Shenyang 110819, China (e-mail: wangxianpeng@ise.neu.edu.cn).}%
}

\maketitle

\begin{abstract}
Intrinsically disordered regions of proteins play a crucial role in cell signaling and drug discovery. However, their high structural flexibility makes accurate residue-level prediction challenging. Existing methods often rely on single-view representations or rigid manual fusion strategies, which fail to effectively balance the complex interplay between local amino acid preferences and long-range sequence patterns. To address these limitations, we propose D2MOE, a Dual-View Multiscale Features and Multi-objective Evolutionary Algorithm, which consists of two stages. First, a dual-view multiscale feature extraction method is introduced. This method integrates evolutionary views with deep semantic views and employs multiscale extractors to capture structural information across diverse receptive fields. Second, a multi-objective evolutionary algorithm is designed to adaptively discover optimal fusion architectures. By co-evolving discrete feature selection and continuous fusion weights, the algorithm adaptively explores optimal cross-feature architectures to enhance predictive accuracy while maintaining model compactness. Experimental results across three benchmark datasets demonstrate that D2MOE consistently outperforms state-of-the-art methods. D2MOE combines the feature extraction capabilities of deep learning with the global search advantages of evolutionary algorithms, enabling efficient feature integration without manual design, and providing a robust computational tool for protein disorder prediction.
\end{abstract}


\section{Introduction}
Intrinsically disordered regions (IDRs) lack a single, stable three-dimensional structure under physiological conditions and instead populate heterogeneous conformational ensembles~\cite{IDRClass2014,IDRFunction2016}. This structural plasticity enables context-dependent recognition and regulation in signalling and transcription, and has been linked to human disease and higher-order cellular organization~\cite{IDRDisease2014,Wu2025_PhaseIDRPred}. Accurate IDRs prediction is therefore important for functional annotation and for interpreting the effects of proteome-wide sequence variation.

Experimental techniques provide accurate IDR characterization, however they are inherently costly and low-throughput~\cite{IDRClass2014,IDRFunction2016}. The exponential growth of sequence databases has vastly outpaced these experimental capabilities, creating a gap between available sequences and structural annotations~\cite{UniProt2023,Conte2023_CAID2}. As a result, computational predictors have become indispensable tools for large-scale IDR identification and proteome analysis~\cite{Kombo2024_AI_IDR}.

Early computational approaches predominantly relied on physics-based energy estimation and machine learning classifiers utilizing hand-crafted features. Energy-based methods, exemplified by the IUPred series, quantify disorder propensity via pairwise interaction potentials, with IUPred3~\cite{IUPred3_2021} further integrating experimental annotations and AIUPred~\cite{AIUPred2024} coupling this energy-centric view with deep learning. In parallel, sequence-based machine learning predictors evolved from integrating evolutionary information to constructing complex, multi-level sequence profiles. A notable example is flDPnn~\cite{flDPnn2020}, which maps high-dimensional, complementary sequence descriptors to disorder scores through a deep feed-forward network. However, these methods remain limited by their reliance on expert-designed features or fixed sliding windows, which may fail to capture long-range dependencies and global context essential for complex disorder modeling~\cite{PONDR1997,IDPpred2019_Biomol}.

Recent advances have shifted IDRs prediction toward deep learning and protein language models (PLMs). Large pre-trained PLMs learn residue-level contextual semantics via self-supervision, thereby enabling a semantic view for disorder modeling~\cite{ProtTrans2021_ProtT5,ESM1b2021_LM}. Current PLM-based predictors primarily rely on ESM and ProtTrans. Regarding methods utilizing ESM embeddings, NetSurfP-3.0~\cite{netsurfp30} adopts ESM residue embeddings as semantic inputs and applies a lightweight CNN--BiLSTM multitask backbone to jointly predict local structural properties, such as secondary structure, disorder, and solvent accessibility. ADOPT~\cite{ADOPT2023_TransformerIDR} trains an ESM-based bidirectional Transformer under the supervision of NMR chemical-shift Z-scores. In comparison, other approaches leverage the ProtTrans series. For instance, LMDisorder~\cite{LMDisorder2023_Ensemble} employs ProtTrans embeddings with a Transformer-based predictor. DisoFLAG~\cite{DisoFLAG2024} combines ProtT5 embeddings with GRU-based sequence modeling and a graph convolutional head to jointly predict disorder states and disorder-related functions. These PLM-based models demonstrate superior accuracy compared to prior methods.

Despite these advances, most predictors rely on single-view and single-scale feature extraction, which limits the richness of the representation and hinders the capture of both short and long disordered domains. Furthermore, the fusion of views and features is often overlooked, leading to inadequacies in selecting and integrating them for information complementarity. Given the complexity of biological information, diverse views and features may exhibit inherent intersections or complementarities. However, current models typically rely on manual feature selection and fusion, which is not only time-consuming and labor-intensive but also hinders effective information utilization and limits prediction performance. Evolutionary algorithms has shown promise for automating feature subset selection and fusion; however, most efforts limited to specific domains, such as image classification and other bioinformatics tasks~\cite{Xiong2025_MultiPop_DE,qianEnhancedProteinSecondary2025}. Moreover, unlike standardized image data, protein data are highly complex, and residue-level IDR prediction poses substantially different challenges.

To address the aforementioned challenges we propose D2MOE, a dual-view multiscale feature and multi-objective evolutionary algorithm, for enhancing protein intrinsic disorder prediction.
The dual-view multiscale feature strategy is designed to extract complementary representations and capture the sequence patterns of both short and long disordered regions. Additionally, the multi-objective evolutionary algorithm is introduced to adaptively perform feature selection and fusion, thereby reducing model complexity and improving prediction performance. The main contributions of this work are as follows:
\begin{enumerate}
    \item A dual-view multiscale feature strategy is developed for intrinsic disorder prediction. The dual-view approach integrates HMM profiles to extract evolutionary information of proteins and ProtT5 embeddings to capture protein semantics, yielding complementary information. RNNs and multiscale CNNs capture global and local features, enriching protein representations and enhancing prediction performance.
    \item A multi-objective evolutionary algorithm (MOEA) is introduced to adaptively perform feature selection and fusion. A differential evolution (DE) strategy introduces optimizable fusion weights, enhancing fusion flexibility. MOEA simultaneously maximizes predictive performance and minimizes the number of selected features, thereby improving predictive accuracy while reducing model complexity.   
    \item D2MOE consistently outperforms baselines across multiple evaluation metrics on three benchmark datasets. Dual-view and multiscale features are superior to alternative designs, and MOEA fusion exceeds fixed manual rules and its single-objective variant, with DE-based weight refinement providing additional gains.
\end{enumerate}

The paper is organized as follows. Section~II reviews related disorder predictors and motivates the proposed design. Section~III describes the proposed framework and methodological details. Section~IV introduces the datasets, implementation details, and metrics. Section~V presents the main results and ablations. Finally, Section~VI concludes the paper.

\section{Related Work}
\subsection{Dual-View Multiscale Features Strategy}

Multi-view and multi-feature learning have improved representation quality and predictive accuracy in vision and NLP, and are increasingly being adopted in bioinformatics~\cite{Chen2024_MFTrans,Wei2023_ConPep}. For protein tasks, multi-view frameworks have been developed for subcellular localization~\cite{Zhang2021_MVSubcell}, drug--target interaction via multi-graph encoders~\cite{Zhang2024_DTI_MV}, and site prediction through the fusion of language-model embeddings with sequence descriptors~\cite{Guan2024_DeepKlapred}. However, these methods typically focus on modeling relationships across different views, while largely overlooking that each view itself can be represented by multiple heterogeneous features, thereby limiting multi-feature integration and adaptive feature selection within each view.

Meanwhile, multiscale feature extraction has also been shown to be valuable for modeling complex biomolecular systems. This is particularly relevant to IDRs, where short irregular segments and long disordered domains co-occur and require both local and long-range receptive fields. Multiscale convolutions and hierarchical extractors have been used to capture patterns across receptive-field scales in drug--target interaction prediction~\cite{Peng2025_MSCMLCIDTI}, antigen--antibody binding modeling~\cite{Li2025_TransABseq}, and protein model-quality assessment~\cite{Shuvo2023_iQDeep}. Collectively, these studies demonstrate that multiscale convolutions and hierarchical feature extractors are effective in capturing structural and interaction patterns across different spatial ranges. These methods motivate the application of multi-view, multiscale feature strategies to intrinsic disorder prediction; however, existing approaches often remain constrained by inadequate feature selection and inflexible fusion. These limitations further motivate adaptive cross-feature fusion strategies for intrinsic disorder prediction.

\begin{figure*}[t]
  \centering
  \includegraphics[width=1\textwidth]{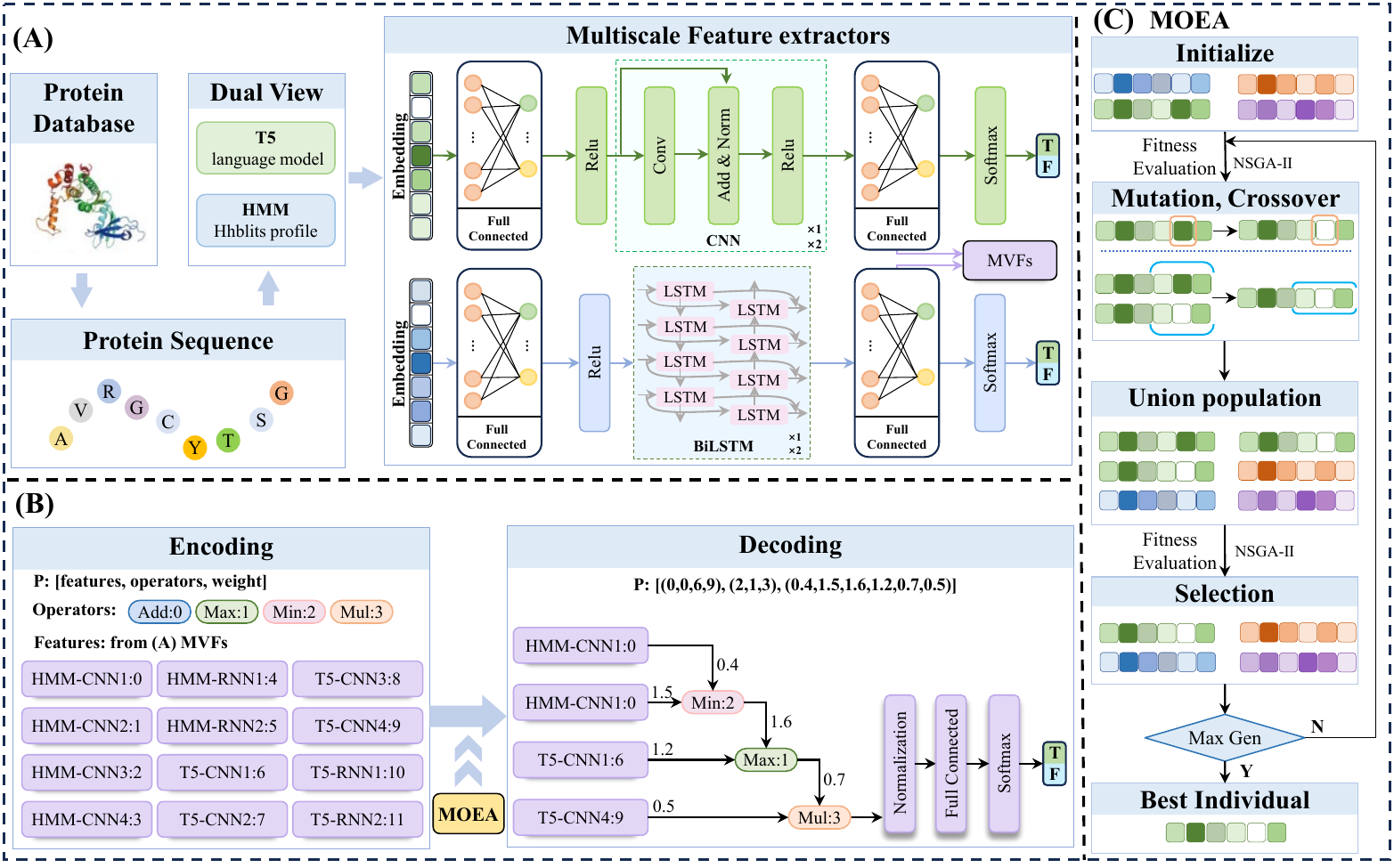}
  \caption{Overview of D2MOE.
\textbf{(A)} Dual-view multiscale feature extraction. Protein sequences are mapped into semantic (ProtT5) and evolutionary (HMM) views, which are then passed through multiscale CNN and RNN extractors to generate complementary feature representations. 
\textbf{(B)} Adaptive feature fusion via \textbf{(C)} MOEA. The 12 candidate features are encoded and evolved using a multi-objective algorithm to simultaneously optimize feature selection, fusion operators (Add, Max, Min, Mul), and fusion weights.}
  \label{fig:all}
\end{figure*}

\subsection{Multi-objective Evolutionary Algorithm}
Information fusion is central to multi-view and multiscale learning because it links heterogeneous feature representations to the final predictor. Most existing approaches are based on hand-crafted, fixed heuristics, such as element-wise maximum or direct concatenation~\cite{Chen2024_MFTrans}. Such designs may not sufficiently model inter-feature dependencies and typically require manual selection of fusion rules, which can limit cross-view interactions and propagate redundant information to downstream predictors~\cite{23,25}. To address these limitations, global optimization methods, including evolutionary algorithms, particle swarm optimization, and genetic programming, have been employed to automatically search feature subsets and fusion structures~\cite{Xiong2025_MultiPop_DE,qianEnhancedProteinSecondary2025}. In particular, the MOEA is commonly adopted to balance predictive performance against model complexity and to obtain compact models with competitive accuracy~\cite{NSGAII2002_Deb,10670082}. Nevertheless, most evolutionary fusion frameworks target generic representations or image-centric pipelines and are seldom tailored to IDRs prediction.

Multi-objective evolutionary feature selection is typically formulated as a discrete subset search that maximizes predictive performance while minimizing the number of selected features~\cite{qianEnhancedProteinSecondary2025}. By contrast, fusion coefficients are continuous variables and are more appropriately optimized in continuous spaces, thereby supporting the coupling of subset selection with continuous weight optimization and the adoption of mixed-variable evolutionary designs with type-specific variation operators for discrete and continuous components~\cite{MMVOP_Review}. DE is a classical optimizer for continuous parameter spaces and has been widely used to tune feature weights and fusion coefficients in biomedical and engineering applications~\cite{DE1997_StornPrice,AmirHosseini2019_FuzzyDE_CT,DE_Survey_FeatureSelection}. However, when the candidate feature set is large, using DE alone to perform both subset selection and weight optimization can substantially expand the search space and exacerbate redundancy. Therefore, coupling multi-objective subset selection with DE-based continuous weight optimization is justified for adaptive, compact, and accurate fusion in dual-view multiscale IDRs prediction, and this NSGA-II--DE co-evolution scheme jointly optimizes the feature subset and fusion weights to improve accuracy while controlling complexity.

\section{Methodology}
The proposed D2MOE framework consists of two stages. In the first stage, a dual-view multiscale feature strategy is adopted to extract protein representations from complementary views and multiscale sequence features. In the second stage, a multi-objective evolutionary algorithm is employed to adaptively perform feature subset selection and weighted fusion, thereby improving predictive performance while reducing model complexity.

\subsection{Dual-View Multiscale Features}

The dual-view multiscale feature strategy comprises two steps: constructing complementary sequence representations and extracting features across multiple receptive-field scales as shown in Fig. \ref{fig:all}(A).

\subsubsection{Dual-View Representation}

IDRs exhibit heterogeneous sequence cues that are difficult to capture using a single information source. DV-MsF therefore represents each residue from two complementary views: an evolutionary view derived from HHblits-based HMM profiles and a semantic view derived from ProtT5 embeddings. These residue-wise representations are used as inputs to the subsequent multiscale feature extractors.

\paragraph{Semantic view from ProtT5 embeddings}
ProtT5 is used to provide a residue-wise semantic representation learned from large-scale self-supervised pretraining, enabling contextual modeling of non-local dependencies~\cite{ProtTrans2021_ProtT5,ESM1b2021_LM}. For a protein sequence of length $L$, ProtT5 yields $E_{\text{T5}} \in \mathbb{R}^{L \times 1024},$ where each row is a 1024-dimensional contextual embedding. Such embeddings have been shown to encode disorder-relevant biochemical and structural cues and to generalize without requiring multiple sequence alignments~\cite{LMDisorder2023_Ensemble,ADOPT2023_TransformerIDR,netsurfp30}.

\paragraph{Evolutionary view from HHblits-HMM profiles}
An evolutionary representation is constructed from HHblits-derived HMM profiles to capture conservation patterns and substitution preferences. For a sequence of length $L$, HHblits produces an $L \times 30$ profile; following common practice, the first 20 substitution-statistics channels are retained as $E_{\text{HMM}} \in \mathbb{R}^{L \times 20}.$ This view provides explicit family-level constraints that complement the contextual semantics encoded by $E_{\text{T5}}$.

\subsubsection{Multiscale Feature Extraction}

The goal of multiscale feature extraction is to improve IDRs prediction performance by capturing a broader range of sequence-derived characteristics. According to the properties of disordered regions and the respective strengths of CNNs and RNNs, six base extractors, CNN1, CNN2, CNN3, CNN4, RNN1, and RNN2, are employed in each view. The formation of IDRs is typically governed by both local amino-acid composition and global structural flexibility, making local and global information equally important. IDRs are governed by both short-range compositional irregularities and long-range conformational flexibility.
Therefore, multiscale CNNs capture local motifs across varying receptive fields, while RNNs integrate sequential context to model global dynamics and long-range dependencies. To accommodate the large variability in IDRs lengths, CNNs with different receptive fields are adopted to cover sequence features at different scales. As summarized in Algorithm~\ref{alg:DV-MSF}, these six extractors jointly extract global and local features from inputs of the two views, and their corresponding architectures are illustrated in Fig.~\ref{fig:all}(A).

All feature extractors follow a unified processing pipeline that includes protein sequence preprocessing, feature extraction, dimensionality reduction, and output generation. Samples from the training set $\mathcal{D}_{\text{tr}}$ and validation set $\mathcal{D}_{\text{val}}$ are used as inputs to train the six extractors CNN1, CNN2, CNN3, CNN4, RNN1, and RNN2 in each view. For a protein sequence of length $L$, the corresponding view input is first concatenated with residue one-hot encodings along the feature dimension to obtain a view-specific sequence representation. This representation is then projected by a fully connected layer with ReLU activation into an $L \times 256$ intermediate feature matrix. The intermediate features are fed into one of the six CNN or RNN extractors to learn multiscale sequence patterns. Subsequently, stacked fully connected layers are applied for channel integration and dimensionality reduction, yielding a 256-dimensional high-level feature vector. The prediction head outputs logits for order and disorder, which are mapped to residue-wise probabilities via a softmax layer. Finally, for each extractor, the 256-dimensional penultimate representation is concatenated with the 2-dimensional logits to form a 258-dimensional candidate feature vector. Applying this pipeline to all six extractors in both views yields 12 candidate descriptors: HMM-CNN1, HMM-CNN2, HMM-CNN3, HMM-CNN4, HMM-RNN1, HMM-RNN2, T5-CNN1, T5-CNN2, T5-CNN3, T5-CNN4, T5-RNN1, and T5-RNN2. These twelve descriptors jointly constitute the candidate feature pool used by the subsequent MOEA-based adaptive fusion module.

\begin{algorithm}[!t]
\caption{Pseudocode of Multiscale Feature Extraction.}
\label{alg:DV-MSF}
\textbf{Input:} Training and Validation dataset with dual views\\
\hspace*{4.5em}$\mathcal{D}_{\text{tr}} = \{(\mathcal{X}_{\text{HMM}}, \mathcal{Y}), (\mathcal{X}_{\text{T5}}, \mathcal{Y})\}$;\\
\hspace*{4.5em}$\mathcal{D}_{\text{val}} = \{(\tilde{\mathcal{X}}_{\text{HMM}}, \tilde{\mathcal{Y}}), (\tilde{\mathcal{X}}_{\text{T5}}, \tilde{\mathcal{Y}})\}$.\\
\textbf{Output:} Multi-feature set $\mathcal{F} = \{\phi_t\}_{t=1}^{12}$.\\[0.2em]
1: Concatenate the view information with a residue one-hot \\
\hspace*{1.2em}encoding to obtain the view embedding;\\
2: Input the view embedding to a fully connected layer with\\
\hspace*{1.2em}ReLU activation, output a $256$-dimensional vector;\\
3: Input the obtained vector to the following extractors:\\
\hspace*{1.8em}- Option 1: One small-kernel CNN layer (CNN1);\\
\hspace*{1.8em}- Option 2: Two small-kernel CNN layers (CNN2);\\
\hspace*{1.8em}- Option 3: One large-kernel CNN layer (CNN3);\\
\hspace*{1.8em}- Option 4: Two large-kernel CNN layers (CNN4);\\
\hspace*{1.8em}- Option 5: One BiLSTM layer (RNN1);\\
\hspace*{1.8em}- Option 6: Two BiLSTM layers (RNN2);\\
4: Input the obtained vector to:\\
\hspace*{1.8em}- Case 1 (CNN1--CNN4): Three fully connected layers\\
\hspace*{1.8em}with ReLU activation, resulting in outputs of 256, 256,\\
\hspace*{1.8em}and 2 dimensions, respectively;\\
\hspace*{1.8em}- Case 2 (RNN1 or RNN2): Three fully connected layers\\
\hspace*{1.8em}with ReLU activation, resulting in outputs of 512, 256,\\
\hspace*{1.8em}and 2 dimensions, respectively;\\
5: Apply the Softmax function to the 2-dimensional logits,\\
\hspace*{1.2em}output residue-wise order or disorder predictions;\\
6: \textbf{return} $\phi_t$ obtained by concatenating the 256-dimensional\\
\hspace*{1.2em}penultimate fully connected layer output and the\\
\hspace*{1.2em}2-dimensional logits, yielding a 258-dimensional feature\\
\hspace*{1.2em}vector.
\end{algorithm}

\subsection{Multi-objective Evolutionary Algorithm}

The MOEA module is employed to adaptively perform feature selection and feature fusion while reducing model complexity and improving predictive accuracy. As shown in Fig.~\ref{fig:all}(B) and (C), the procedure consists of three components: encoding, evolutionary search, and decoding. First, candidate features are encoded into individual representations. NSGA-II is then employed to jointly optimize two objectives, prediction performance and the number of selected features, thereby adaptively selecting feature subsets and determining the fusion structure. Finally, solutions on the Pareto front are decoded to obtain concrete fusion models. In this way, NSGA-II avoids the limitations of manually designed architectures and alleviates the negative effects of feature redundancy, whereas the DE algorithm optimizes fusion weights across features, highlighting informative features and further improving the overall predictive performance of the model.

\subsubsection{Encoding}

Let $\mathcal{F}$ denote the set of 12 candidate features produced by DV--MsF, and let
$\Omega=\{Add, Mul, Max, Min\}$ be the set of basic
element-wise fusion operators, where Add/Mul/Max/Min correspond to weighted
summation,\\ 
Hadamard product, element-wise maximum, and element-wise minimum,
respectively. Let $\mathbf{a}$ be the fusion weight matrix whose elements are aligned
with the corresponding features during fusion. Each individual is represented as $\chi=[\mathbf{s},\mathbf{q},\mathbf{a}],$ where $\mathbf{s}=(s_1,\ldots,s_n)$ is a sequence of selected features from $\mathcal{F}$, and a feature may be selected more than once, $\mathbf{q}=(q_1,\ldots,q_{n-1})$
is the operator sequence, and
$\mathbf{a}=(a_{1,1},a_{1,2},\ldots,a_{n-1,1},a_{n-1,2})$ is the corresponding
fusion-weight vector. The number of features, $n$, is individual-specific with
$n\in[1,n_{\max}]$ and $n_{\max}=12$.

\subsubsection{Decoding}
Given an individual $\chi$, it is decoded into a left-fold fusion tree, as illustrated in Fig.~\ref{fig:all}(B). We first initialize $\mathbf{c}_1=\mathbf{s}_1$. For $t=2,\ldots,n$, we iteratively compute
\begin{equation}
\tilde{\mathbf{c}}_{t-1}=a_{t,1}\ast\mathbf{c}_{t-1},\quad
\tilde{\mathbf{s}}_t=a_{t,2}\ast\mathbf{s}_t,\quad
\mathbf{c}_t=q_t(\tilde{\mathbf{c}}_{t-1},\tilde{\mathbf{s}}_t).
\label{eq:fuse_en}
\end{equation}
The final descriptor $\mathbf{c}_n$ is then mapped to a probability vector
$\hat{y}$ through a fully connected layer whose weight matrix is estimated by
the least-squares method; a subsequent Softmax function is applied to $\hat{y}$
to obtain the final prediction $\tilde{\mathcal{Y}}$.

In this representation, $\mathbf{s}$ and $\mathbf{q}$ jointly determine which
features and operators are selected in the fusion network and how they are
combined. For implementation convenience, we index the 12 candidate features by
integers $0\!\sim\!11$ and map the four operators to integers $0\!\sim\!3$; in
what follows, we still use $s_t$ and $q_t$ to denote the selected feature and
operator, respectively.

\subsubsection{Workflow of the MOEA}

As illustrated in Algorithm~\ref{alg:MOEA}, follows an NSGA-II + DE co-evolution scheme: NSGA-II optimizes \((\mathbf{s},\mathbf{q})\) under a multi-objective selection mechanism, while DE refines \(\mathbf{a}\) in the continuous space. It consists of four major steps: population initialization (line 1), fitness evaluation (lines 3 and 10), mutation and crossover (lines 8–9), and new population generation plus optimal solution selection (lines 13 and 17).

\paragraph{Population initialization}

Given a population size \(N\) and maximal generation \(G_{\max}\), the initial population \(\mathcal{P}^{(0)}\) is randomly generated. Each individual is represented as
\(\chi_k=[\mathbf{s}_k,\mathbf{q}_k,\mathbf{a}_k]\), where the feature number \(n_k=\mathrm{length}(\mathbf{s}_k)\) may vary across individuals.

\paragraph{Fitness evaluation}
The fitness of each individual is evaluated using two objectives: maximizing accuracy and minimizing the feature number. Accuracy is measured by the AUC metric, whereas the feature number is used to reduce the number of selected features, avoid redundancy, and lower model complexity.

\begin{itemize}
  \item Accuracy: For a given individual \(\chi_k\), predictions \(\tilde{\mathcal{Y}}\) are first computed according to the decoding procedure. The AUC is defined as
  \begin{equation}
    F_1(\chi_k)=\mathrm{AUC}\!\left(\{(\hat{y}_j,y_j)\}_{j=1}^{N_{\mathrm{res}}}\right),
    \label{eq:auroc_en}
  \end{equation}
  where \(y_j\in\{0,1\}\) and \(\hat{y}_j\in[0,1]\) denote the ground-truth labels and predicted disorder probabilities, and \(N_{\mathrm{res}}\) is the total number of residues.

  \item Feature number: The second objective aims to minimize redundant features in the fusion tree so as to avoid performance degradation and reduce model complexity. For an individual
  \(\chi_k=[\mathbf{s}_k,\mathbf{q}_k,\mathbf{a}_k]\), the feature number is defined as
  \begin{equation}
    F_2(\chi_k)=n_k=\mathrm{length}(\mathbf{s}_k),
    \label{eq:featnum_en}
  \end{equation}
  where \(n_k\) denotes the number of features used in individual \(\chi_k\).
\end{itemize}

\begin{algorithm}[t]
\caption{Framework of MOEA.}
\label{alg:MOEA}
\textbf{Input:} Candidate feature set $\mathcal{F}$; fusion operator set $\Omega$;\\
\hspace*{3.2em}population size $N$; maximal generation $G_{\max}$;\\
\hspace*{3.2em}labels $\mathcal{Y}$; mutation probability $P_{\text{mu}}$; population $\mathcal{P}$\\
\hspace*{3.2em}maximum number of features in an individual $n_{\max}$.\\
\textbf{Output:} Best individual $\chi_{\text{best}}$ and its corresponding fusion\\
\hspace*{3.5em}network.\\[0.2em]
1: Initialize the population $\mathcal{P}^{(0)} = \{\chi_k=[\mathbf{s}_k,\mathbf{q}_k,\mathbf{a}_k]\}_{k=1}^{N}$\\
\hspace*{1.2em}according to the encoding scheme and set the current\\
\hspace*{1.2em}generation index $g = 0$.\\
2: \textbf{for} each individual $\chi_k$ in $\mathcal{P}^{(0)}$ \textbf{do}\\
3:\hspace*{1.0em} Decode $\chi_k$ into its corresponding fusion network, and\\
\hspace*{2.2em}compute fitness values $F_1(\chi_k)$ and $F_2(\chi_k)$.\\
4: \textbf{end for}\\
5: \textbf{while} $g < G_{\max}$ \textbf{do}\\
6:\hspace*{1.3em} $\mathcal{U} = \mathrm{COPY}(\mathcal{P}^{(g)})$;\\
7:\hspace*{1.3em} \textbf{for} $k = 1$ to $N$ \textbf{do}\\
8:\hspace*{2.0em} $\chi_m = \mathrm{HYBRID\_MUTATION}(\chi_k, P_{\text{mu}}, \mathcal{P}^{(g)})$;\\
9:\hspace*{2.0em} $\chi_c = \mathrm{HYBRID\_CROSSOVER}(\chi_k, \chi_m)$;\\
10:\hspace*{2.0em}Decode $\chi_m$ and $\chi_c$ into their corresponding fusion\\
  \hspace*{3.2em}networks, and compute the fitness of  $\chi_m$ and $\chi_c$;\\
11:\hspace*{2.0em}$\mathcal{U} = \mathcal{U} \cup \{\chi_m\} \cup \{\chi_c\}$;\\
12:\hspace*{1.0em} \textbf{end for}\\
13:\hspace*{1.0em} $\mathcal{P}^{(g+1)} = \mathrm{GET\_NEXT\_GENERATION}(\mathcal{U})$;\hspace*{0.4em}\\
14:\hspace*{1.0em} $\mathcal{U} = \emptyset$;\\
15:\hspace*{1.0em} $g = g + 1$;\hspace*{0.4em}// Update generation index\\
16: \textbf{end while}\\
17: $\chi_{\text{best}} \leftarrow$ Select the individual with the best fitness from\\
\hspace*{1.2em} the final Pareto front.\\
18: \textbf{return} the fusion network decoded from $\chi_{\text{best}}$.
\end{algorithm}

\paragraph{Hybrid Mutation and Crossover}
Hybrid mutation and crossover are adopted in MOEA. Hybrid mutation simultaneously modifies the integer part of the chromosome (feature selection $\mathbf{s}$ and operator selection $\mathbf{q}$) and the real-valued part (fusion weight matrix $\mathbf{a}$). Integer mutation randomly perturbs selected features or operators, whereas real-valued mutation updates the weights using the DE strategy. Hybrid crossover performs single-point crossover between parent chromosomes by exchanging segments of features, operators, and weights to generate offspring.

The detailed procedures of hybrid mutation and hybrid crossover used in Algorithm~\ref{alg:MOEA} are given below.

i) Hybrid Mutation. Since an individual $\chi_k$ contains both an integer part ($\mathbf{s}_k$ and $\mathbf{q}_k$) and a real-valued part ($\mathbf{a}_k$), different mutation strategies are applied to generate a new chromosome vector
\(\chi_m=[\mathbf{s}_m,\mathbf{q}_m,\mathbf{a}_m]\), where $\chi_m$ is initialized as a copy of $\chi_k$, and $n_m=\mathrm{length}(\mathbf{s}_m)$ denotes the current number of selected features.

Integer Component Mutation:
For the feature sequence \(\mathbf{s}_m\) and operator sequence \(\mathbf{q}_m\), integer mutation consists of three operations:
\begin{itemize}
  \item Generate a random number \(R\). If \(R\le P_{\text{mu}}\), resample \(n\sim U(1,n_{\max})\).
  If \(n<n_m\), truncate \(\mathbf{s}_m,\mathbf{q}_m,\mathbf{a}_m\) to length $n$; if \(n>n_m\), extend \(\mathbf{s}_m\) with random features from $\mathcal{F}$, extend \(\mathbf{q}_m\) with random operators from $\Omega$, and initialize the newly added weights uniformly in \([0.1,2]\).
  \item Generate another random number \(R\). If \(R\le P_{\text{mu}}\), randomly select an index \(t\in[1,n_m]\) and replace the corresponding feature $s_m[t]$ with another candidate in $\mathcal{F}$.
  \item Generate another random number \(R\). If \(R\le P_{\text{mu}}\), randomly select an index \(j\in[1,n_m-1]\) and replace the corresponding operator $q_m[j]$ with an operator in $\Omega$.
\end{itemize}

Real-Valued Component Mutation:
To update the continuous fusion weights, an improved DE-based mutation is adopted:
\begin{equation}
\mathbf{a}_m
=
\mathbf{a}_m
+\rho(\mathbf{a}^1-\mathbf{a}_m)
+\rho(\mathbf{a}^2-\mathbf{a}^3),
\label{eq:de_en}
\end{equation}
where $\mathbf{a}^1,\mathbf{a}^2,\mathbf{a}^3$ are the real-valued parts of three individuals \(\chi^1,\chi^2,\chi^3\), respectively. The three individuals \(\chi^1,\chi^2,\chi^3\) are randomly sampled from the population \(\mathcal{P}^{(0)}\) and sorted in descending order according to $F_1$. The differential weight decays linearly with the generation index $g$:
\begin{equation}
\rho=\rho_0\Bigl(1-\frac{g}{G_{\max}}\Bigr),\quad \rho_0=0.9.
\label{eq:rho_en}
\end{equation}

ii) Crossover Operation. Crossover is applied between two parent individuals \(\chi_k=[\mathbf{s}_k,\mathbf{q}_k,\mathbf{a}_k]\) and \(\chi_m=[\mathbf{s}_m,\mathbf{q}_m,\mathbf{a}_m]\) to generate a new chromosome vector \(\chi_c=[\mathbf{s}_c,\mathbf{q}_c,\mathbf{a}_c]\).

\emph{Step 1:} Randomly select a crossover point \(n_c\) such that \(n_c \in \{1,\ldots,\min(n_k,n_m)\}\).

\emph{Step 2:} Perform synchronized single-point crossover on the three components:
\begin{equation}
\begin{aligned}
\mathbf{s}_c &= \mathrm{concat}(\mathbf{s}_k[1:n_c-1],\,\mathbf{s}_m[n_c:n_m]),\\
\mathbf{q}_c &= \mathrm{concat}(\mathbf{q}_k[1:n_c-1],\,\mathbf{q}_m[n_c:n_m-1]),\\
\mathbf{a}_c &= \mathrm{concat}(\mathbf{a}_k[1:n_c-1],\,\mathbf{a}_m[n_c:n_m-1]).
\end{aligned}
\label{eq:cross_en}
\end{equation}
The resulting $\chi_c$ is taken as the offspring produced by hybrid crossover.

\paragraph{Generation of Offspring and Selection}
At each generation, the parent, mutated, and crossover populations are merged into a joint pool \(\mathcal{U}\). Under the two objectives of maximizing predictive accuracy and minimizing feature number, NSGA-II performs fast non-dominated sorting and crowding-distance based selection, retaining the top \(N\) individuals to form the next generation. After reaching \(G_{\max}\) or satisfying a convergence criterion, the solution with the highest fitness score on the final Pareto front is selected as \(\chi_{\text{best}}\), which encodes the final feature subset, operator sequence, and weight configuration for compact and accurate IDRs prediction.

\section{Experiments}

\subsection{Datasets}

The training and validation protocol of NetSurfP\mbox{-}3.0 was followed~\cite{netsurfp30}, and model performance was evaluated on three established benchmark test sets:

\begin{itemize}
  \item \textbf{TS115}~\cite{TS115}: Consisting of 115 non-redundant PDB proteins with low sequence similarity, TS115 is widely used for stringent evaluation, as reduced redundancy and structural diversity provide a more demanding test of robustness.
  \item \textbf{CASP12}~\cite{CASP12}: Originating from the CASP12 blind assessment (21 targets), CASP12 contains proteins of varying complexity and thereby better reflects realistic test conditions, offering a challenging scenario for IDRs prediction.
  \item \textbf{CB513}~\cite{CB513}: Comprising 513 segments from 434 proteins, CB513 covers diverse structural and functional contexts and serves as a standard benchmark for assessing generalization in residue-level IDRs prediction.
\end{itemize}

For efficient mini-batch training, each protein sequence was standardized to a maximum length of 700 residues. Sequences shorter than 700 residues were padded, while longer sequences were split into 700-residue segments. To prevent padded positions from affecting the training objective, we applied loss masking, and this padding-aware strategy aligns with prior work~\cite{netsurfp30}. Specifically, we used a PDB-derived non-redundant split with standard filtering: proteins shorter than 20 residues were excluded and clusters containing validation sequences were removed, yielding 10,837 proteins; 500 were randomly selected for validation and 10,337 for training.

\subsection{Experimental Setting}
All experiments were run on a single NVIDIA GeForce RTX~3090 GPU with 24~GB memory, using Python~3.9 and PyTorch~2.0.0 with CUDA~11.8. The Adam optimizer implemented in PyTorch was used for training. The training was stopped when the validation loss did not improve for 10 epochs, and the checkpoint with the lowest validation loss across all epochs was selected as the final model. Such early stopping was used to avoid potential overfitting on smaller datasets. All hyperparameters were selected through empirical testing and established best practices to optimize performance and ensure training stability. Table~\ref{tab:hyperparams} summarizes the key hyperparameters used in this study.

\begin{table}[!t]
\caption{Hyperparameters used in D2MOE.}
\label{tab:hyperparams}
\centering
\renewcommand{\arraystretch}{1.15}
\small
\begin{tabular*}{\linewidth}{@{\extracolsep{\fill}} @{\hspace{2.5em}} l c @{\hspace{3.0em}}}
\toprule
\textbf{Hyperparameter} & \textbf{Setting} \\
\midrule
Number of epochs & 100 \\
Dropout rate & 0.2 \\
Batch size & 256 \\
Maximum sequence length & 700 \\
Number of labels & 2 \\
Kernel sizes (multiscale CNNs) & 9, 17 \\
Padding lengths (CNNs) & 4, 8 \\
Number of channels (CNNs) & 256 \\
Hidden layer size(RNNs) & 256 \\
Fully connected layers (CNNs) & 256, 256, 2 \\
Fully connected layers (BiLSTM) & 512, 256, 2 \\
Maximum generations (MOEA) & 100 \\
Population size (MOEA) & 40 \\
Union population size (MOEA) & 120 \\
Crossover probability (MOEA) & 0.4 \\
Mutation probability (MOEA) & 0.25 \\
Mutation control factor range & [0.1, 0.9] \\
\bottomrule
\end{tabular*}
\end{table}

\begin{figure*}[!t]
  \centering

  \begin{minipage}{0.5\textwidth}
    \centering
    \begin{overpic}[width=\linewidth]{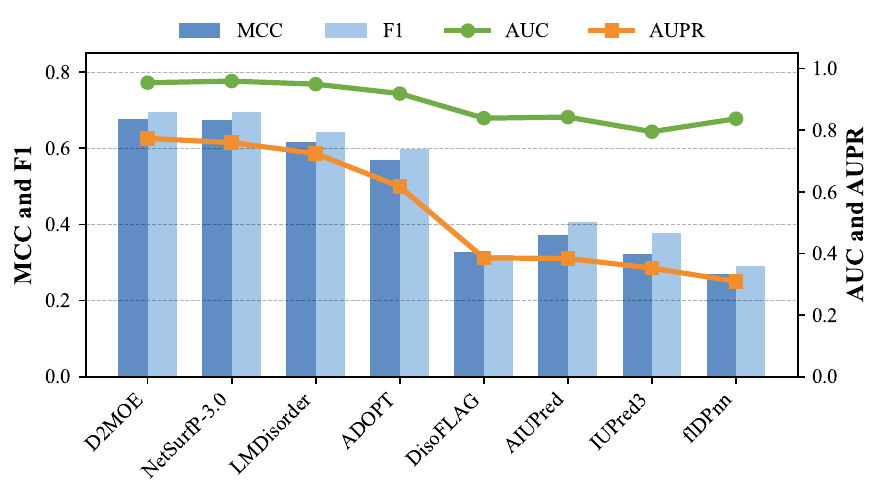}
      \put(1,53){\large a}
    \end{overpic}
  \end{minipage}%
  \begin{minipage}{0.5\textwidth}
    \centering
    \begin{overpic}[width=\linewidth]{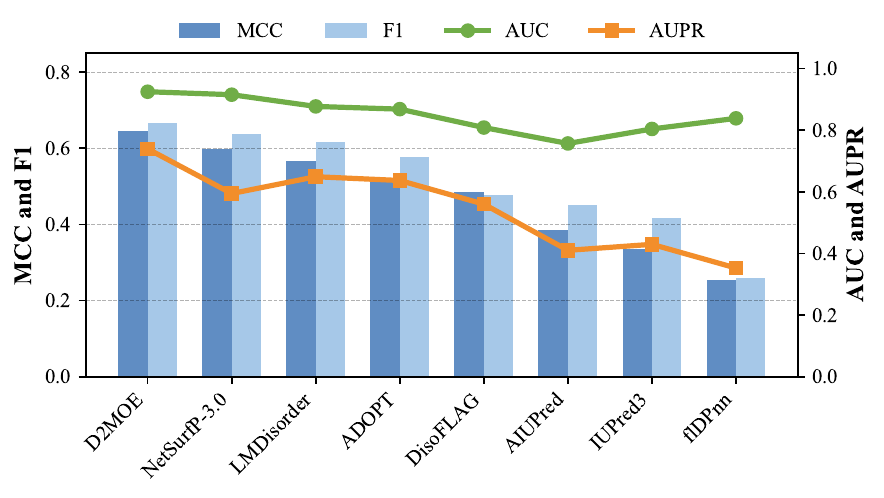}
      \put(1,53){\large b}
    \end{overpic}
  \end{minipage}
  
  \vspace{1em} %

  \begin{minipage}{0.5\textwidth}
    \centering
    \begin{overpic}[width=\linewidth]{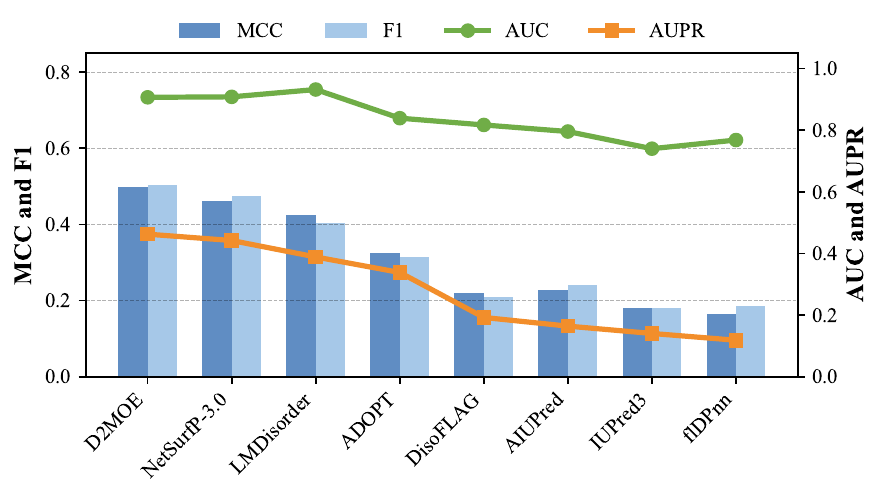}
      \put(1,53){\large c}
      \end{overpic}
  \end{minipage}%
  \begin{minipage}{0.5\textwidth}

    \vspace*{-2em}
    
    \begin{minipage}{0.5\linewidth}
      \centering
      \begin{overpic}[width=\linewidth]{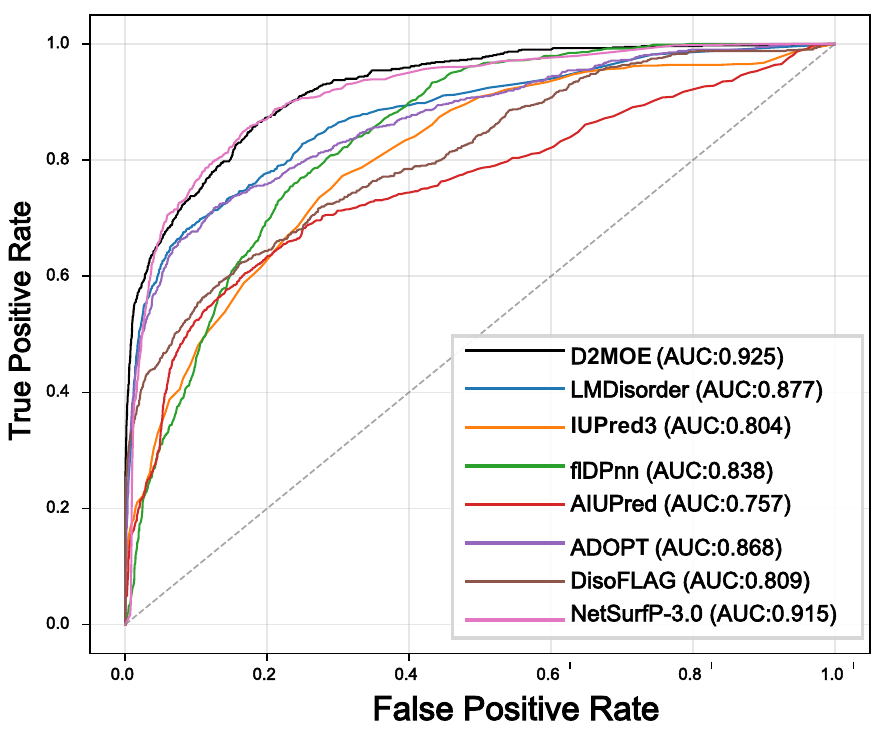}
        \put(1,85){\large d}
      \end{overpic}
    \end{minipage}%
    \begin{minipage}{0.5\linewidth}
      \centering
      \begin{overpic}[width=\linewidth]{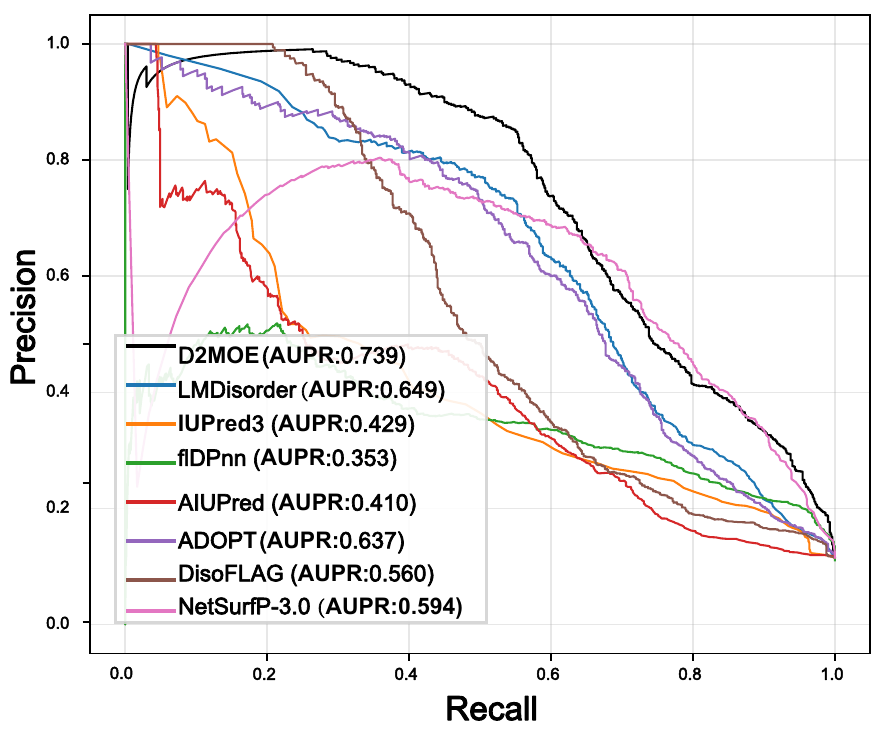}
        \put(1,85){\large e}
      \end{overpic}
    \end{minipage}
    
  \end{minipage}
  
  \vspace{0.5em}
  \caption{Performance comparison of D2MOE and seven representative predictors on three benchmark datasets. 
  (a--c) Quantitative results on TS115, CASP12 and CB513, reporting MCC and F1 (bars) together with AUC and AUPR (lines). 
  (d,e) ROC and PR curves on CASP12; AUC and AUPR are reported in the legends.}
  \label{fig:ablation_combined}
\end{figure*}

\subsection{Evaluation Metrics}
Model performance was evaluated using four metrics: Matthews Correlation Coefficient (MCC), area under the ROC curve (AUC), area under the Precision--Recall curve (AUPR), and F1. MCC and F1 were computed on binarized predictions using a fixed threshold $\tau=0.5$ (a residue is predicted as disordered if its propensity $\ge \tau$).

\begin{equation}
\mathrm{MCC} =
\frac{\mathrm{TP}\cdot \mathrm{TN} - \mathrm{FP}\cdot \mathrm{FN}}
{\sqrt{(\mathrm{TP}+\mathrm{FP})(\mathrm{TP}+\mathrm{FN})(\mathrm{TN}+\mathrm{FP})(\mathrm{TN}+\mathrm{FN})}}
\end{equation}
where $\mathrm{TP}$, $\mathrm{TN}$, $\mathrm{FP}$, and $\mathrm{FN}$ are the numbers of true positives, true negatives, false positives, and false negatives, respectively.

\begin{equation}
\mathrm{AUPR} = \int_0^1 \mathrm{Precision}\!\left(\mathrm{Recall}^{-1}(u)\right)\, du
\end{equation}
where Precision reflects the reliability of predicted disordered residues and Recall reflects the coverage of true disordered residues.

\begin{equation}
\mathrm{AUC} = \int_0^1 \mathrm{TPR}\!\left(\mathrm{FPR}^{-1}(u)\right)\, du
\end{equation}
where TPR captures sensitivity to disordered residues and FPR captures the propensity to misclassify ordered residues as disordered.

\begin{equation}
\mathrm{F1}=\frac{2\cdot \mathrm{Precision}\cdot \mathrm{Recall}}{\mathrm{Precision}+\mathrm{Recall}}
\end{equation}
where all symbols follow the definitions above.

\section{Results}
The results are presented in four parts: first, D2MOE is benchmarked against seven representative IDRs predictors on CB513, TS115, and CASP12; second, the impact of the dual-view design is evaluated; third, the benefit of multiscale and hybrid feature extraction is examined; and finally, the effectiveness of the proposed multi-objective evolutionary fusion is assessed.

\subsection{Comparison With State-of the-Art Methods}

D2MOE was benchmarked against seven representative IDRs predictors on CB513, TS115, and CASP12, including energy-based methods IUPred3~\cite{IUPred3_2021} and AIUPred~\cite{AIUPred2024}, a profile and descriptor integration model flDPnn~\cite{flDPnn2020}, and four PLM-driven predictors NetSurfP\mbox{-}3.0~\cite{netsurfp30}, LMDisorder~\cite{LMDisorder2023_Ensemble}, DisoFLAG~\cite{DisoFLAG2024}, and ADOPT~\cite{ADOPT2023_TransformerIDR}. Results are summarized in Fig.~\ref{fig:ablation_combined}(a--c), with CASP12 ROC and PR curves in Fig.~\ref{fig:ablation_combined}(d,e).  

D2MOE ranks first in MCC and AUPR on all three benchmarks. The gain is most pronounced on CASP12, improving MCC by 7.9\% over NetSurfP\mbox{-}3.0 and AUPR by 13.9\% over LMDisorder, consistent with the ROC and PR curves in Fig.~\ref{fig:ablation_combined}(d,e). On TS115, MCC and AUPR reach 0.677 and 0.773, while AUC and F1 remain close to NetSurfP\mbox{-}3.0; on CB513, MCC and AUPR increase by 7.6\% and 4.7\% over NetSurfP\mbox{-}3.0. Notably, LMDisorder attains the highest AUC on CB513, yet its lower MCC and AUPR indicate that improved ranking does not necessarily yield a favourable PR trade-off under imbalance. More broadly, PLM-based predictors NetSurfP\mbox{-}3.0, LMDisorder, DisoFLAG, and ADOPT generally outperform energy-based methods IUPred3 and AIUPred and the profile and descriptor integration model flDPnn in MCC and AUPR, highlighting the value of contextual semantic representations. Building on this advantage, D2MOE augments ProtT5-derived semantics with an explicit evolutionary view from HHblits-HMM profiles and exploits complementary evidence across multiple receptive-field scales and long-range dependencies. In addition, MOEA-based adaptive fusion searches for compact, discriminative feature combinations and calibrates cross-view contributions, which can suppress redundancy and stabilize PR performance. Collectively, these results suggest that combining PLM semantics with complementary evolutionary constraints and adaptive fusion yields more reliable residue-level disorder prediction across diverse benchmarks.

\begin{figure*}[htbp]
  \centering
  \begin{minipage}{0.333\textwidth}  
    \centering
    \includegraphics[width=\linewidth]{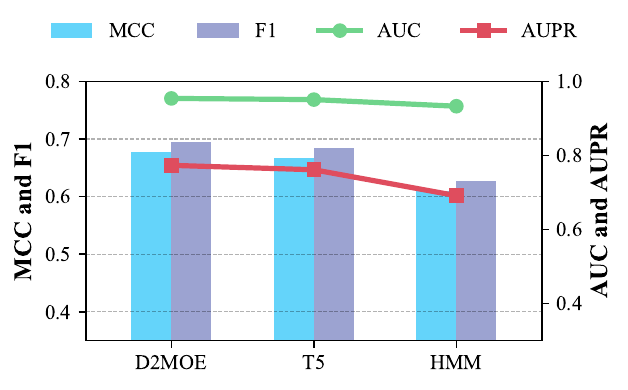}
    
    \vspace{0.25em}
    {\footnotesize (a) TS115}
  \end{minipage}%
  \hfill
  \begin{minipage}{0.333\textwidth}  
    \centering
    \includegraphics[width=\linewidth]{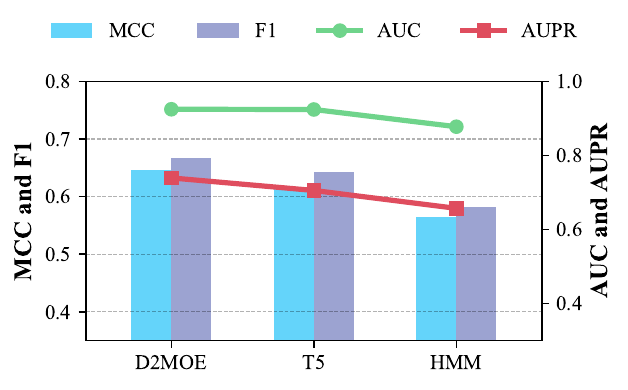}
    
    \vspace{0.25em}
    {\footnotesize (b) CASP12}
  \end{minipage}%
  \hfill
  \begin{minipage}{0.333\textwidth}  %
    \centering
    \includegraphics[width=\linewidth]{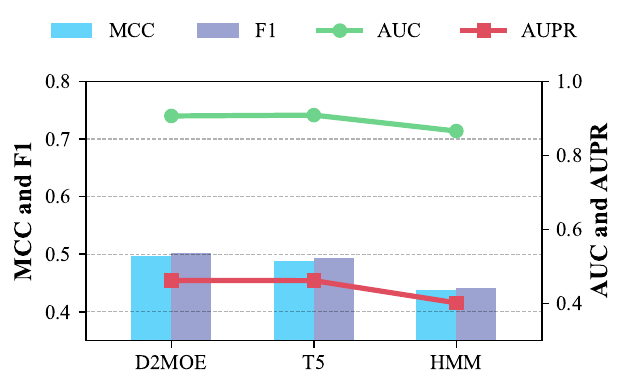}
    
    \vspace{0.25em}
    {\footnotesize (c) CB513}
  \end{minipage}%

\caption{Validation of the dual-view design on TS115, CASP12 and CB513. D2MOE denotes the dual-view model that integrates ProtT5 embeddings with HHblits-HMM profiles, whereas T5 and HMM use the semantic view or the evolutionary view alone. Bars report MCC and F1 (left axis), and lines report AUC and AUPR (right axis).}
  \label{fig:dualview_ablation3}
\end{figure*}

\begin{figure*}[htbp]
  \centering
  \begin{minipage}{0.333\textwidth}  
    \centering
    \includegraphics[width=\linewidth]{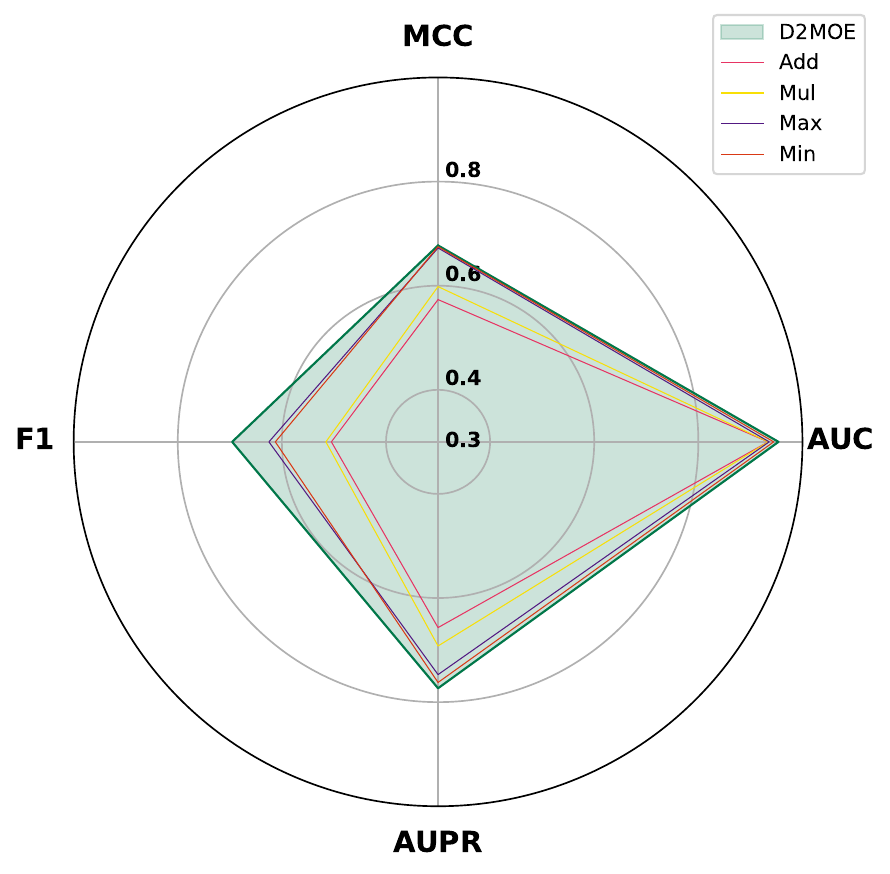}
    
    \vspace{0.25em}
    {\footnotesize (a) TS115}
  \end{minipage}%
  \hfill
  \begin{minipage}{0.333\textwidth}  %
    \centering
    \includegraphics[width=\linewidth]{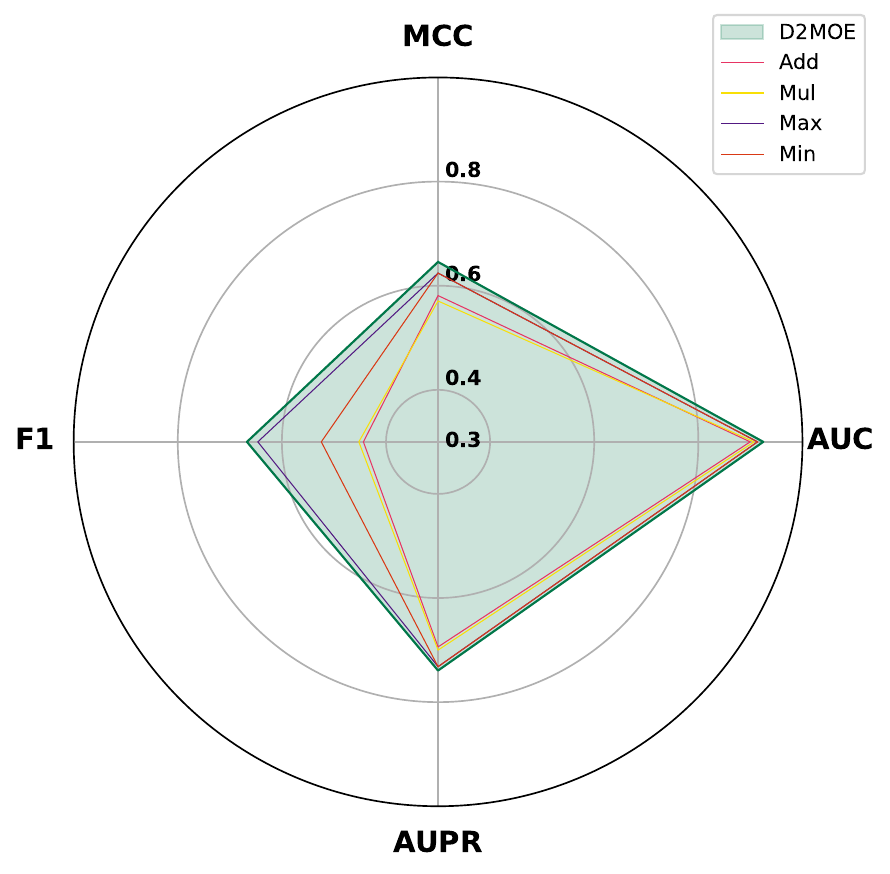}
    
    \vspace{0.25em}
    {\footnotesize (b) CASP12}
  \end{minipage}%
  \hfill
  \begin{minipage}{0.333\textwidth}  %
    \centering
    \includegraphics[width=\linewidth]{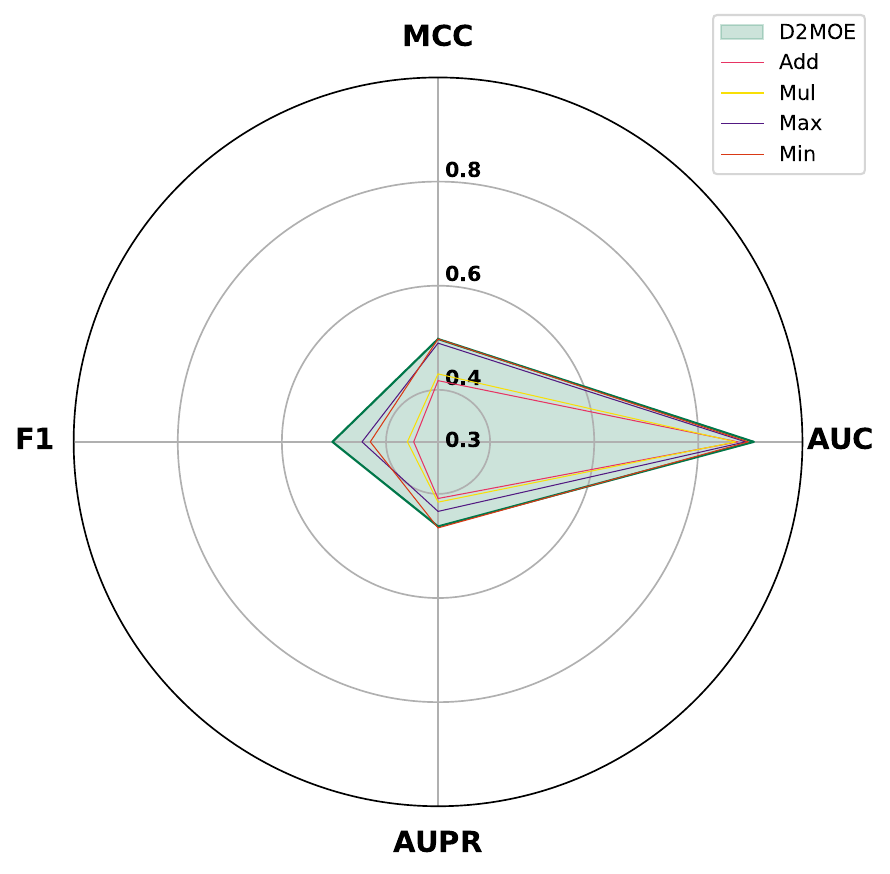}
    
    \vspace{0.25em}
    {\footnotesize (c) CB513}
  \end{minipage}%

\caption{Radar comparison of five fusion schemes (Add, Mul, Max, Min, and D2MOE) on (a) TS115, (b) CASP12, and (c) CB513 in terms of MCC, AUC, AUPR, and F1 (larger radius indicates better performance). D2MOE encloses all fixed operators on TS115 and CASP12, and remains dominant on CB513 except that Min is slightly higher on AUPR.}

  \label{fig:ablation_three_datasets}
\end{figure*}

\subsection{Validation of the Dual-view Design}
To assess whether jointly modeling semantic and evolutionary evidence benefits residue-level disorder prediction, the dual-view D2MOE was compared with two single-view variants, T5 and HMM, on CB513, TS115 and CASP12 (Fig.~\ref{fig:dualview_ablation3}). D2MOE performs best on TS115 and CASP12, achieving the highest MCC and AUPR while remaining competitive in AUC and F1. In particular, on TS115, D2MOE attains an MCC of 0.677 and an AUPR of 0.773, exceeding T5; on CASP12, it reaches an MCC of 0.646 and an AUPR of 0.739, again outperforming T5. On CB513, D2MOE yields the best MCC and F1, with an AUPR comparable to T5 and a marginally lower AUC. Overall, T5 consistently outperforms HMM across all four metrics, indicating that ProtT5-derived semantic representations provide stronger discriminative evidence than evolutionary statistics alone. Moreover, the additional gains obtained by D2MOE over T5 on TS115 and CASP12 support the complementarity between ProtT5 semantics and HHblits-HMM evolutionary profiles, and show that their integration leads to more accurate and stable disorder prediction across benchmarks.

\subsection{Validation of Multiscale and Hybrid Feature Extraction}
To evaluate the contribution of multiscale convolutions, D2MOE was compared with two single-scale hybrid variants, DV-SSF and DV-BSF, in Table~\ref{tab:multiscale_ablation}. DV-SSF combines small-kernel convolutions with an RNN module, whereas DV-BSF adopts large-kernel convolutions while keeping the same hybrid backbone. On CB513, D2MOE achieves the best MCC and F1, while AUC and AUPR are slightly lower than those of the single-scale hybrids. More importantly, D2MOE achieves the best overall performance on TS115, reaching an AUC of 0.954, an AUPR of 0.773, an MCC of 0.677, and an F1 of 0.696. On CASP12, D2MOE also achieves the best overall performance, attaining an MCC of 0.646 and an AUPR of 0.739. These results suggest that integrating multiple receptive fields better accommodates disordered segments of varying lengths, improves feature expressiveness, and improves prediction performance.

To assess the effect of hybrid CNN and RNN modeling, D2MOE was further compared with two constrained variants that use a single feature extractor, CNN-only and RNN-only, under otherwise identical settings in Table~\ref{tab:multiscale_ablation}. On CB513, D2MOE again yields the highest MCC and F1, with an MCC of 0.498 and an F1 of 0.503, while AUC is comparable and AUPR is lower than those of the single-extractor variants. On TS115 and CASP12, the hybrid model improves all four metrics over both single-extractor variants. Encouragingly, on CASP12, the MCC increases by approximately 4.5\% relative to the CNN-only model, rising from 0.618 to 0.646. Overall, coupling CNN-based local pattern extraction with RNN-based long-range context modeling provides richer representations and yields more consistent gains in residue-level disorder prediction.

\begin{table*}[!t]
\caption{Ablation experiments for multiscale feature strategy on TS115, CASP12, and CB513}
\label{tab:multiscale_ablation}
\centering
\renewcommand{\arraystretch}{1.17}
\resizebox{\textwidth}{!}{
\begin{tabular}{@{}l|cccc|cccc|cccc@{}}
\toprule
Datasets & \multicolumn{4}{c|}{TS115} & \multicolumn{4}{c|}{CASP12} & \multicolumn{4}{c}{CB513} \\
\midrule
Methods & MCC & AUPR & F1 & AUC & MCC & AUPR & F1 & AUC & MCC & AUPR & F1 & AUC \\
\midrule
RNNs-only     & 0.656 & 0.739 & 0.674 & 0.940 & 0.622 & 0.721 & 0.648 & 0.924 & 0.484 & 0.477 & 0.489 & 0.910 \\
CNNs-only     & 0.671 & 0.749 & 0.690 & 0.951 & 0.618 & 0.721 & 0.646 & 0.917 & 0.492 & \textbf{0.483} & 0.498 & 0.906 \\
DV-SSF        & 0.669 & 0.756 & 0.687 & 0.953 & 0.605 & 0.714 & 0.633 & 0.917 & 0.494 & 0.476 & 0.495 & 0.907 \\
DV-BSF        & 0.666 & 0.754 & 0.682 & 0.951 & 0.606 & 0.721 & 0.634 & 0.919 & 0.488 & 0.466 & 0.490 & \textbf{0.916} \\
D2MOE   & \textbf{0.677} & \textbf{0.773} & \textbf{0.696} & \textbf{0.954} & \textbf{0.646} & \textbf{0.739} & \textbf{0.667} & \textbf{0.925} & \textbf{0.498} & 0.463 & \textbf{0.503} & 0.907 \\
\bottomrule
\end{tabular}
}
\end{table*}

\begin{table*}[!t]
\caption{PERFORMANCE COMPARISON WITH DIFFERENTIAL EVOLUTION COEFFICIENTS}
\label{tab:performance_w1_vs_dvmsfmoea}
\centering
\resizebox{\textwidth}{!}{
\begin{tabular}{@{}l|cccc|cccc|cccc@{}}
\toprule
Datasets & \multicolumn{4}{c|}{TS115} & \multicolumn{4}{c|}{CASP12} & \multicolumn{4}{c}{CB513} \\
\midrule
Methods & MCC & AUPR & F1 & AUC & MCC & AUPR & F1 & AUC & MCC & AUPR & F1 & AUC \\
\midrule
W=1          & 0.674 & 0.748 & 0.690 & 0.947 & 0.632 & 0.710 & 0.657 & 0.909 & 0.496 & 0.452 & 0.500 & 0.898 \\
D2MOE  & \textbf{0.677} & \textbf{0.773} & \textbf{0.696} & \textbf{0.954} & \textbf{0.646} & \textbf{0.739} & \textbf{0.667} & \textbf{0.925} & \textbf{0.498} & \textbf{0.463} & \textbf{0.503} & \textbf{0.907} \\
\bottomrule
\end{tabular}
}
\end{table*}

\subsection{Evaluation of Multi-objective Evolutionary Algorithm}
The contribution of the MOEA is examined from three aspects: comparison against fixed element-wise fusion operators, the effect of DE-based fusion weights, and the multi-objective formulation that enforces compact yet complementary feature integration beyond any single descriptor.

\subsubsection{Comparison With Fixed Fusion Operators}
To assess whether the MOEA-derived fusion is preferable to hand-crafted element-wise rules, it was compared with four fixed operators (Add, Mul, Max, and Min). As shown in Fig.~\ref{fig:ablation_three_datasets}(a--c), D2MOE forms the outermost radar profile on TS115 and CASP12, fully enclosing the contours of all four fixed operators, indicating consistent advantages across all four metrics. On CB513, D2MOE likewise provides the dominant envelope over most metric dimensions. These results underscore that effective fusion is critical when exploiting the multi-view multiscale feature library, and that automated MOEA-driven fusion is generally preferable to fixed, manually specified element-wise operators.

\subsubsection{Effect of DE Weight Optimization}
To evaluate the contribution of DE weight optimization, two independent MOEA searches were performed. In the DE-enabled setting, fusion coefficients were evolved in the continuous space, whereas in the weight-free variant all fusion coefficients were fixed to unity (denoted as W=1). Each setting was assessed using the best individual returned by its corresponding search. As summarized in Table~\ref{tab:performance_w1_vs_dvmsfmoea}, enabling DE yields consistently stronger performance across CB513, TS115, and CASP12. These results indicate that learning data-adaptive fusion weights helps calibrate the relative contributions of the selected features, providing greater fusion flexibility than uniform weighting and translating into more reliable predictive quality across benchmarks.

\subsubsection{Multi-objective Feature Selection and Adaptive Feature Fusion}
To assess the value of the multi-objective formulation, the D2MOE was compared with the single-objective DV-MsF-EA, which optimises predictive performance alone and does not penalise feature-set size. Table~\ref{tab:performance_f2_vs_dvmsfmoea} shows that the two variants are broadly comparable on the three datasets, whereas D2MOE shows stronger generalization on TS115 and CASP12. On TS115, D2MOE achieves MCC of 0.677 and F1 of 0.696, compared with MCC of 0.675 and F1 of 0.694 for DV-MsF-EA. On CASP12, D2MOE reaches MCC of 0.646 and F1 of 0.667, compared with MCC of 0.623 and F1 of 0.654. Importantly, these gains are obtained with a substantially more compact solution: the selected D2MOE uses seven features, whereas the selected DV-MsF-EA uses twelve, indicating that the multi-objective formulation can reduce redundancy while preserving accuracy.

\begin{figure}
  \includegraphics[width=\columnwidth]{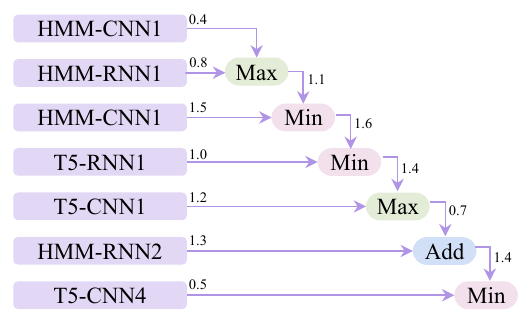}
  \caption{The best fused architecture using MOEA.}
  \label{fig:BEST}
\end{figure}

The best-evolved fusion architecture in Fig.~\ref{fig:BEST} helps explain why a compact solution remains competitive. The selected model integrates semantic and evolutionary representations and combines multiscale local cues with longer-range dependencies, rather than relying on a single view or a single extractor family. This interpretation is supported by the feature-level evidence in Table~\ref{tab:12vsdvmsfmoea}. Some single features attain strong threshold-independent scores on CB513, for example, T5-CNN1 reaches AUC of 0.918, yet none delivers uniformly strong threshold-based performance across benchmarks.
On TS115, the best single feature reaches MCC of 0.669 and F1 of 0.689, remaining below D2MOE with MCC of 0.677 and F1 of 0.696. CASP12 shows the same trend. Collectively, the gains are best explained by complementary cross-view and cross-scale signals captured through adaptive fusion, rather than by any single descriptor or by expanding the feature set.

\begin{table*}[ht]
\caption{Evaluation of multi-objective (D2MOE) and single-objective (DV-MsF-EA) evolutionary fusion on TS115, CASP12, and CB513.}
\label{tab:performance_f2_vs_dvmsfmoea}
\centering
\resizebox{\textwidth}{!}{
\begin{tabular}{@{}l|cccc|cccc|cccc@{}}
\toprule
Datasets & \multicolumn{4}{c|}{TS115} & \multicolumn{4}{c|}{CASP12} & \multicolumn{4}{c}{CB513} \\
\midrule
Methods & MCC & AUPR & F1 & AUC & MCC & AUPR & F1 & AUC & MCC & AUPR & F1 & AUC \\
\midrule
DV-MsF-EA    & 0.675 & 0.773 & 0.694 & 0.953 & 0.623 & 0.735 & 0.654 & 0.917 & \textbf{0.499} & \textbf{0.483} & \textbf{0.505} & 0.905 \\
D2MOE  & \textbf{0.677} & \textbf{0.773} & \textbf{0.696} & \textbf{0.954} & \textbf{0.646} & \textbf{0.739} & \textbf{0.667} & \textbf{0.925} & 0.498 & 0.463 & 0.503 & \textbf{0.907} \\
\bottomrule
\end{tabular}
}
\end{table*}

\begin{table*}[ht]
\caption{Performance of the D2MOE model and twelve single features on TS115, CASP12, and CB513.}
\label{tab:12vsdvmsfmoea}
\centering
\resizebox{\textwidth}{!}{
\begin{tabular}{@{}l|cccc|cccc|cccc@{}}
\toprule
Datasets & \multicolumn{4}{c|}{TS115} & \multicolumn{4}{c|}{CASP12} & \multicolumn{4}{c}{CB513} \\
\midrule
Methods & MCC & AUPR & F1 & AUC & MCC & AUPR & F1 & AUC & MCC & AUPR & F1 & AUC \\
\midrule
HMM-CNN1     & 0.528 & 0.623 & 0.518 & 0.912 & 0.411 & 0.534 & 0.408 & 0.841 & 0.371 & 0.338 & 0.356 & 0.841 \\
HMM-CNN2     & 0.606 & 0.692 & 0.618 & 0.932 & 0.539 & 0.624 & 0.554 & 0.869 & 0.424 & 0.389 & 0.429 & 0.857 \\
HMM-CNN3     & 0.578 & 0.679 & 0.572 & 0.927 & 0.492 & 0.619 & 0.494 & 0.868 & 0.404 & 0.380 & 0.396 & 0.855 \\
HMM-CNN4     & 0.614 & 0.693 & 0.634 & 0.935 & 0.585 & 0.669 & 0.605 & 0.871 & 0.434 & 0.405 & 0.442 & 0.869 \\
T5-CNN1      & 0.663 & 0.763 & 0.679 & 0.950 & 0.624 & 0.736 & 0.647 & 0.923 & 0.486 & 0.469 & 0.487 & \textbf{0.918} \\
T5-CNN2      & 0.669 & 0.771 & 0.685 & 0.952 & 0.619 & 0.739 & 0.647 & 0.923 & 0.493 & \textbf{0.471} & 0.494 & 0.913 \\
T5-CNN3      & 0.666 & 0.762 & 0.689 & 0.949 & 0.633 & 0.735 & 0.660 & 0.916 & 0.488 & 0.460 & 0.498 & 0.910 \\
T5-CNN4      & 0.664 & 0.766 & 0.681 & 0.950 & 0.626 & 0.733 & 0.640 & 0.912 & 0.483 & 0.457 & 0.483 & 0.905 \\
HMM-RNN1     & 0.589 & 0.678 & 0.610 & 0.937 & 0.566 & 0.709 & 0.586 & 0.906 & 0.431 & 0.413 & 0.436 & 0.874 \\ 
HMM-RNN2     & 0.574 & 0.680 & 0.596 & 0.936 & 0.561 & 0.691 & 0.575 & 0.904 & 0.416 & 0.383 & 0.421 & 0.869 \\
T5-RNN1      & 0.668 & 0.751 & 0.687 & 0.942 & 0.618 & 0.733 & 0.646 & 0.922 & 0.486 & 0.450 & 0.491 & 0.897 \\
T5-RNN2      & 0.658 & 0.759 & 0.668 & 0.951 & 0.594 & 0.734 & 0.621 & 0.921 & 0.476 & 0.467 & 0.468 & 0.917 \\
D2MOE  & \textbf{0.677} & \textbf{0.773} & \textbf{0.696} & \textbf{0.954} & \textbf{0.646} & \textbf{0.739} & \textbf{0.667} & \textbf{0.925} & \textbf{0.498} & 0.463 & \textbf{0.503} & 0.907 \\
\bottomrule
\end{tabular}
}
\end{table*}

\section{Conclusion}
This study introduces D2MOE, a Dual-View Multiscale Features and Multi-objective Evolutionary framework for enhancing residue-level intrinsic disorder prediction. The framework integrates two distinct sequence views and employs multiscale CNNs to capture local patterns at varying lengths, together with an RNN to encode long-range dependencies, thereby enriching protein representations. Furthermore, a multi-objective evolutionary search is employed to select informative feature subsets and automatically design cross-feature fusion architectures, reducing the rigidity and redundancy of handcrafted fusion rules. Additionally, DE is used to optimize fusion weights for the selected subset and fusion structure, thereby further improving performance. Overall, the multi-objective formulation better balances predictive performance and compactness by reducing model complexity while maintaining predictive quality. Experiments on three benchmark test sets demonstrate that D2MOE outperforms state-of-the-art methods over multiple evaluation metrics while eliminating dependence on manually specified fusion architectures. 

Despite these strengths, limitations remain. First, while the evolved feature subsets and fusion architectures are promising, their interpretation remains challenging, limiting decision traceability and biological insights. Future work will focus on enhancing model interpretability. Second, although the framework performs well on standard tasks, it faces challenges in more complex ones, such as Protein Functional Site Prediction. Future research will explore incorporating topological data and additional features to improve generalization across tasks.

\section*{Acknowledgements}
This work was supported in part by the National Natural Science Foundation of China (NSFC) under Grant Nos. 62522309, and in part by the Natural Science Foundation of Liaoning Province under Grant No. 2025JH6/101000012.



\begin{thebibliography}{10}
\providecommand{\url}[1]{#1}
\csname url@samestyle\endcsname
\providecommand{\newblock}{\relax}
\providecommand{\bibinfo}[2]{#2}
\providecommand{\BIBentrySTDinterwordspacing}{\spaceskip=0pt\relax}
\providecommand{\BIBentryALTinterwordstretchfactor}{4}
\providecommand{\BIBentryALTinterwordspacing}{\spaceskip=\fontdimen2\font plus
\BIBentryALTinterwordstretchfactor\fontdimen3\font minus \fontdimen4\font\relax}
\providecommand{\BIBforeignlanguage}[2]{{%
\expandafter\ifx\csname l@#1\endcsname\relax
\typeout{** WARNING: IEEEtran.bst: No hyphenation pattern has been}%
\typeout{** loaded for the language `#1'. Using the pattern for}%
\typeout{** the default language instead.}%
\else
\language=\csname l@#1\endcsname
\fi
#2}}
\providecommand{\BIBdecl}{\relax}
\BIBdecl

\bibitem{IDRClass2014}
R.~{van der Lee}, M.~Buljan, B.~Lang \emph{et~al.}, ``Classification of intrinsically disordered regions and proteins,'' \emph{Chemical Reviews}, vol. 114, no.~13, pp. 6589--6631, Jul. 2014.

\bibitem{IDRFunction2016}
M.~M. Babu, ``The contribution of intrinsically disordered regions to protein function, cellular complexity, and human disease,'' \emph{Biochemical Society Transactions}, vol.~44, no.~5, pp. 1185--1200, Oct. 2016.

\bibitem{IDRDisease2014}
V.~N. Uversky, ``The triple power of d{$^3$}: Protein intrinsic disorder in degenerative diseases,'' \emph{Frontiers in Bioscience (Landmark Edition)}, vol.~19, no.~2, pp. 181--258, Jan. 2014.

\bibitem{Wu2025_PhaseIDRPred}
X.~Wu, K.~Wang, G.~Hu \emph{et~al.}, ``Empirical assessment of sequence-based predictions of intrinsically disordered regions involved in phase separation,'' \emph{Biomolecules}, vol.~15, no.~8, p. 1079, Jul. 2025.

\bibitem{UniProt2023}
{UniProt Consortium}, ``Uniprot: The universal protein knowledgebase in 2023,'' \emph{Nucleic Acids Research}, vol.~51, no.~D1, pp. D523--D531, Jan. 2023.

\bibitem{Conte2023_CAID2}
A.~D. Conte, M.~Mehdiabadi, A.~Bouhraoua \emph{et~al.}, ``Critical assessment of protein intrinsic disorder prediction ( {\textsc{caid}} ) - results of round 2,'' \emph{Proteins: Structure, Function, and Bioinformatics}, vol.~91, no.~12, pp. 1925--1934, Dec. 2023.

\bibitem{Kombo2024_AI_IDR}
D.~C. Kombo, M.~J. LaMarche, C.~C. Konkankit \emph{et~al.}, ``Application of artificial intelligence and machine learning techniques to the analysis of dynamic protein sequences,'' \emph{Proteins: Structure, Function, and Bioinformatics}, vol.~92, no.~10, pp. 1234--1241, Oct. 2024.

\bibitem{IUPred3_2021}
G.~Erd{\H o}s, M.~Pajkos, and Z.~Doszt{\'a}nyi, ``Iupred3: Prediction of protein disorder enhanced with unambiguous experimental annotation and visualization of evolutionary conservation,'' \emph{Nucleic Acids Research}, vol.~49, no.~W1, pp. W297--W303, Jul. 2021.

\bibitem{AIUPred2024}
G.~Erd{\H o}s and Z.~Doszt{\'a}nyi, ``Aiupred: Combining energy estimation with deep learning for the enhanced prediction of protein disorder,'' \emph{Nucleic Acids Research}, vol.~52, no.~W1, pp. W176--W181, Jul. 2024.

\bibitem{flDPnn2020}
G.~Hu, A.~Katuwawala, K.~Wang \emph{et~al.}, ``fldpnn: Accurate intrinsic disorder prediction with putative propensities of disorder functions,'' \emph{Nature Communications}, vol.~12, no.~1, p. 4438, Jul. 2021.

\bibitem{PONDR1997}
P.~Romero, Z.~Obradovic, C.~Kissinger \emph{et~al.}, ``Identifying disordered regions in proteins from amino acid sequence,'' in \emph{Proceedings of International Conference on Neural Networks (ICNN'97)}, vol.~1, Jun. 1997, pp. 90--95 vol.1.

\bibitem{IDPpred2019_Biomol}
D.~Chaurasiya, R.~Mondal, T.~Lahiri \emph{et~al.}, ``Idppred: A new sequence-based predictor for identification of intrinsically disordered protein with enhanced accuracy,'' \emph{Journal of Biomolecular Structure and Dynamics}, vol.~43, no.~2, pp. 957--965, Jan. 2025.

\bibitem{ProtTrans2021_ProtT5}
A.~Elnaggar, M.~Heinzinger, C.~Dallago \emph{et~al.}, ``Prottrans: Toward understanding the language of life through self-supervised learning,'' \emph{IEEE Transactions on Pattern Analysis and Machine Intelligence}, vol.~44, no.~10, pp. 7112--7127, 2022.

\bibitem{ESM1b2021_LM}
A.~Rives, J.~Meier, T.~Sercu \emph{et~al.}, ``Biological structure and function emerge from scaling unsupervised learning to 250 million protein sequences,'' \emph{Proceedings of the National Academy of Sciences}, vol. 118, no.~15, p. e2016239118, 2021.

\bibitem{netsurfp30}
M.~H. H{\o}ie, E.~N. Kiehl, B.~Petersen \emph{et~al.}, ``Netsurfp-3.0: Accurate and fast prediction of protein structural features by protein language models and deep learning,'' \emph{Nucleic Acids Research}, vol.~50, no.~W1, pp. W510--W515, Jul. 2022.

\bibitem{ADOPT2023_TransformerIDR}
I.~Redl, C.~Fisicaro, O.~Dutton \emph{et~al.}, ``Adopt: Intrinsic protein disorder prediction through deep bidirectional transformers,'' \emph{NAR Genomics and Bioinformatics}, vol.~5, no.~2, p. lqad041, Mar. 2023.

\bibitem{LMDisorder2023_Ensemble}
Y.~Song, Q.~Yuan, S.~Chen \emph{et~al.}, ``Fast and accurate protein intrinsic disorder prediction by using a pretrained language model,'' \emph{Briefings in Bioinformatics}, vol.~24, no.~4, p. bbad173, May 2023.

\bibitem{DisoFLAG2024}
Y.~Pang and B.~Liu, ``Disoflag: accurate prediction of protein intrinsic disorder and its functions using graph-based interaction protein language model,'' \emph{BMC biology}, vol.~22, no.~1, p.~3, 2024.

\bibitem{Xiong2025_MultiPop_DE}
Y.~Xiong, S.~Li, J.~He \emph{et~al.}, ``A prior information-based multi-population multi-objective optimization for estimating 18f-fdg pet/ct pharmacokinetics of hepatocellular carcinoma,'' \emph{BMC Medical Imaging}, vol.~25, no.~1, p.~59, Feb. 2025.

\bibitem{qianEnhancedProteinSecondary2025}
Y.~Qian, L.~Su, M.~Xu \emph{et~al.}, ``Enhanced protein secondary structure prediction through multi-view multi-feature evolutionary deep fusion method,'' \emph{IEEE Transactions on Emerging Topics in Computational Intelligence}, vol.~9, no.~5, pp. 3352--3363, Oct. 2025.

\bibitem{Chen2024_MFTrans}
Y.~Chen, G.~Chen, and C.~Y.-C. Chen, ``Mftrans: A multi-feature transformer network for protein secondary structure prediction,'' \emph{International Journal of Biological Macromolecules}, vol. 267, p. 131311, May 2024.

\bibitem{Wei2023_ConPep}
Q.~Wei, R.~Wang, Y.~Jiang \emph{et~al.}, ``Conpep: Prediction of peptide contact maps with pre-trained biological language model and multi-view feature extracting strategy,'' \emph{Computers in Biology and Medicine}, vol. 167, p. 107631, Dec. 2023.

\bibitem{Zhang2021_MVSubcell}
Q.~Zhang, Y.~Zhang, S.~Li \emph{et~al.}, ``Accurate prediction of multi-label protein subcellular localization through multi-view feature learning with rbrl classifier,'' \emph{Briefings in Bioinformatics}, p. bbab012, Feb. 2021.

\bibitem{Zhang2024_DTI_MV}
Z.~Zhang, X.~He, D.~Long \emph{et~al.}, ``Enhancing generalizability and performance in drug--target interaction identification by integrating pharmacophore and pre-trained models,'' \emph{Bioinformatics}, vol.~40, no. Supplement\_1, pp. i539--i547, Jun. 2024.

\bibitem{Guan2024_DeepKlapred}
J.~Guan, P.~Xie, D.~Dong \emph{et~al.}, ``Deepklapred: A deep learning framework for identifying protein lysine lactylation sites via multi-view feature fusion,'' \emph{International Journal of Biological Macromolecules}, vol. 283, p. 137668, Dec. 2024.

\bibitem{Peng2025_MSCMLCIDTI}
J.~Peng, X.~Liu, Y.~Liao \emph{et~al.}, ``Mscmlcidti: Drug--target interaction prediction based on multiscale feature extraction and deep interactive attention fusion mechanisms,'' \emph{Journal of Computational Chemistry}, vol.~46, no.~19, p. e70170, Jul. 2025.

\bibitem{Li2025_TransABseq}
C.-F. Li, Z.~Yan, F.~Ge \emph{et~al.}, ``Transabseq: A two-stage approach for predicting antigen--antibody binding affinity changes upon mutation based on protein sequences,'' \emph{Journal of Chemical Information and Modeling}, vol.~65, no.~10, pp. 5188--5204, May 2025.

\bibitem{Shuvo2023_iQDeep}
M.~H. Shuvo, M.~Karim, and D.~Bhattacharya, ``iqdeep: An integrated web server for protein scoring using multiscale deep learning models,'' \emph{Journal of Molecular Biology}, vol. 435, no.~14, p. 168057, Jul. 2023.

\bibitem{23}
X.~Wang, T.~Hu, and L.~Tang, ``A multiobjective evolutionary nonlinear ensemble learning with evolutionary feature selection for silicon prediction in blast furnace,'' \emph{IEEE Transactions on Neural Networks and Learning Systems}, vol.~33, no.~5, pp. 2080--2093, 2022.

\bibitem{25}
X.~Wang, Y.~Wang, L.~Tang \emph{et~al.}, ``Multiobjective ensemble learning with multiscale data for product quality prediction in iron and steel industry,'' \emph{IEEE Transactions on Evolutionary Computation}, vol.~28, no.~4, pp. 1099--1113, 2024.

\bibitem{NSGAII2002_Deb}
K.~Deb, A.~Pratap, S.~Agarwal \emph{et~al.}, ``A fast and elitist multiobjective genetic algorithm: Nsga-ii,'' \emph{IEEE Transactions on Evolutionary Computation}, vol.~6, no.~2, pp. 182--197, Apr. 2002.

\bibitem{10670082}
H.~Peng, Z.~Luo, T.~Fang \emph{et~al.}, ``Micro many-objective evolutionary algorithm with knowledge transfer,'' \emph{IEEE Transactions on Emerging Topics in Computational Intelligence}, vol.~9, no.~1, pp. 43--56, 2025.

\bibitem{MMVOP_Review}
E.-G. Talbi, ``Metaheuristics for variable-size mixed optimization problems: A unified taxonomy and survey,'' \emph{Swarm and Evolutionary Computation}, vol.~89, p. 101642, 2024.

\bibitem{DE1997_StornPrice}
R.~Storn and K.~Price, ``Differential evolution -- a simple and efficient heuristic for global optimization over continuous spaces,'' \emph{Journal of Global Optimization}, vol.~11, no.~4, pp. 341--359, Dec. 1997.

\bibitem{AmirHosseini2019_FuzzyDE_CT}
B.~AmirHosseini and R.~Hosseini, ``An improved fuzzy-differential evolution approach applied to classification of tumors in liver ct scan images,'' \emph{Medical \& Biological Engineering \& Computing}, vol.~57, no.~10, pp. 2277--2287, Oct. 2019.

\bibitem{DE_Survey_FeatureSelection}
M.~F. Ahmad, N.~A.~M. Isa, W.~H. Lim \emph{et~al.}, ``Differential evolution: A recent review based on state-of-the-art works,'' \emph{Alexandria Engineering Journal}, vol.~61, no.~5, pp. 3831--3872, 2022.

\bibitem{TS115}
Y.~Yang, J.~Gao, J.~Wang \emph{et~al.}, ``Sixty-five years of the long march in protein secondary structure prediction: the final stretch?'' \emph{Brief Bioinform}, vol.~19, no.~3, pp. 482--494, 12 2016.

\bibitem{CASP12}
L.~A. Abriata, G.~E. Tamò, B.~Monastyrskyy \emph{et~al.}, ``Assessment of hard target modeling in casp12 reveals an emerging role of alignment-based contact prediction methods,'' \emph{Proteins}, vol.~86, no.~S1, pp. 97--112, 2018.

\bibitem{CB513}
J.~A. Cuff and G.~J. Barton, ``Evaluation and improvement of multiple sequence methods for protein secondary structure prediction,'' \emph{Proteins}, vol.~34, no.~4, pp. 508--519, 1999.

\end{thebibliography}
\end{document}